\documentclass{article}

\usepackage{microtype}
\usepackage{graphicx}
\usepackage{subfig}
\usepackage{booktabs} 

\usepackage{comment}
\usepackage{amsmath}
\def \submission {}
\usepackage{color}
\usepackage{ulem}
\usepackage{multirow}
\usepackage{float}
\usepackage{caption}
\usepackage{array}
\usepackage{arydshln}
\usepackage{adjustbox}

\definecolor{Crimson}{rgb}{0.86, 0.08, 0.24}
\definecolor{DarkGreen}{rgb}{0.00, 0.60, 0.00}
\definecolor{RoyalBlue}{rgb}{0.15, 0.25, 0.54}
\definecolor{DarkCyan}{rgb}{0.0, 0.54, 0.54}
\ifx \submission \undefined

\else

\fi

\usepackage{bbm}
\usepackage{bm}

\usepackage{booktabs}       
\usepackage{amsfonts}       
\usepackage{nicefrac}       
\usepackage{microtype}      
\usepackage{color}
\usepackage{subfig}

\usepackage{amsthm}
\usepackage[para,online,flushleft]{threeparttable}
\usepackage{lineno}

\newcommand{\bp}{{\mathbf{p}}}

\newcommand{\bx}{{\mathbf{x}}}

\newcommand{\bA}{{\mathbf{A}}}

\newcommand{\bI}{{\mathbf{I}}}

\newcommand{\bbR}{{\mathbb{R}}}

\newcommand{\cB}{{\mathcal{B}}}
\newcommand{\cD}{{\mathcal{D}}}

\newcommand{\cL}{{\mathcal{L}}}

\newcommand{\ie}{\textit{i}.\textit{e}.}
\newcommand{\eg}{\textit{e}.\textit{g}.}

\def\argmin{\mathop{\mathrm{arg}\, \mathrm{min}}\limits}

\def\argmin{\mathop{\mathrm{arg}\, \mathrm{min}}\limits}

\usepackage{hyperref}

\usepackage[accepted]{icml2021}

\icmltitlerunning{Submission and Formatting Instructions for ICML 2021}

\begin{document}

\twocolumn[
\icmltitle{ACE: Adaptive Confusion Energy for Natural World Data Distribution}



\icmlsetsymbol{equal}{*}

\begin{icmlauthorlist}
\icmlauthor{Yen-Chi Hsu}{iis, ntu}
\icmlauthor{Cheng-Yao Hong}{iis}
\icmlauthor{Wan-Cyuan Fan}{ntu}
\icmlauthor{Ming-Sui Lee}{ntu}
\icmlauthor{Davi Geiger}{nyu}
\icmlauthor{Tyng-Luh Liu}{iis}
\end{icmlauthorlist}

\icmlaffiliation{iis}{Academia Sinica}
\icmlaffiliation{ntu}{National Taiwan University}
\icmlaffiliation{nyu}{New York University}

\icmlcorrespondingauthor{Yen-Chi Hsu}{d06922021@csie.ntu.edu.tw}
\icmlcorrespondingauthor{Cheng-Yao Hong}{sensible@iis.sinica.edu.tw}

\icmlkeywords{Machine Learning, ICML}

\vskip 0.3in
]




\begin{abstract}
With the development of deep learning, standard classification problems have achieved good results. However, conventional classification problems are often too idealistic. Most data in the natural world usually have imbalanced distribution and fine-grained characteristics. Recently, many state-of-the-art approaches tend to focus on one or another separately, but rarely on both. In this paper, we introduce a novel and adaptive batch-wise regularization based on the proposed Adaptive Confusion Energy (ACE) to flexibly address the nature world distribution, which usually involves fine-grained and long-tailed properties at the same time. ACE increases the difficulty of the training process and further alleviates the overfitting problem. Through the datasets with the technical issue in fine-grained (CUB, CAR, AIR) and long-tailed (ImageNet-LT), or comprehensive issues (CUB-LT, iNaturalist), the result shows that the ACE is not only competitive to some state-of-the-art on performance but also demonstrates the effectiveness of training.
\end{abstract}

%
\section{Introduction}
\label{sec:intro}
%

With the development of deep learning, the fundamental classification problem has been solved. Subsequent classification studies focus on two more challenging issues, fine-grained characteristics, and imbalanced data distribution. Fine-grained visual classification (FGVC) is an active and challenging problem in computer vision. Such a recognition task differs from the classical problem of large-scale visual classification (LSVC) by focusing on differentiating {\em similar} sub-categories of the same meta-category. In FGVC, the inter-class similarity among the object categories is often pervasive. The intra-class variations further impose ambiguities in learning a unified and discriminative representation for each category. Long-tailed distribution brings another aspect of the challenge that the head categories tend to dominate the training procedure. Thus, the learned classification model performs better on these categories while yielding significantly poor performance for the tail categories. The performance distribution somewhat resembles the data distribution. As the natural world distribution often assumes both fine-grained and long-tailed properties, how to satisfactorily address the recognition problem under such a general setting raises a practical and challenging issue.

\begin{figure}[t!]
    \centering
    \includegraphics[width=0.48\textwidth]{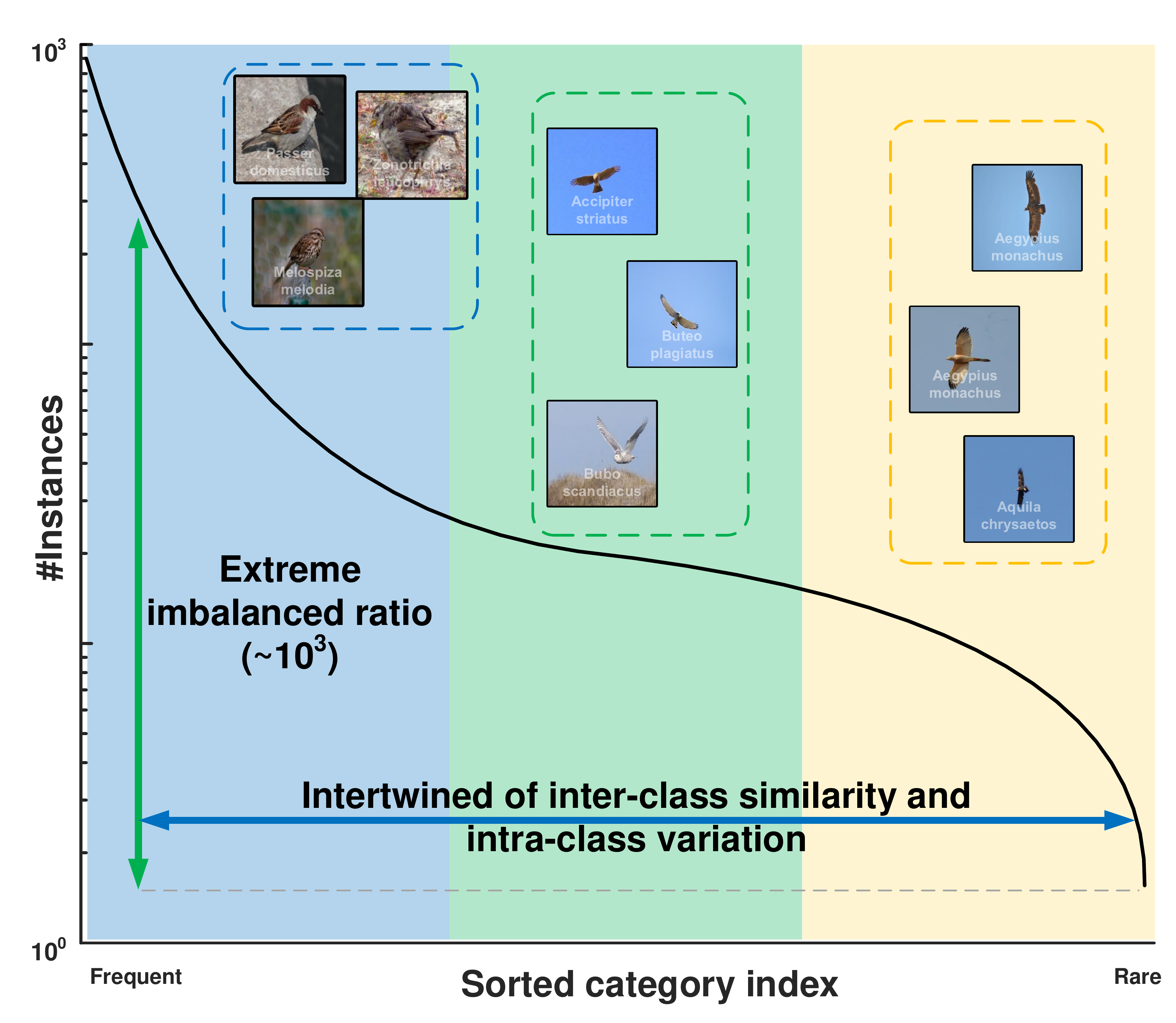}
    \caption{Data distribution is not ideal in the real world and is usually accompanied by more than one complicated issue to be solved. For example, iNaturalist 2018~\cite{van2018inaturalist} hard to learn the tailed classes due to an extreme imbalanced ratio in the \textit{long-tailed} distribution. Meanwhile, it is hard to disentangle the inter-class similarity and intra-class variation that results from \textit{fine-grained} characteristics.  Eventually, this leads to overfitting.}
    \label{fig:intro}
    \vspace{-5pt}
\end{figure}

\begin{figure*}[ht!]
    \centering
    \includegraphics[width=\textwidth]{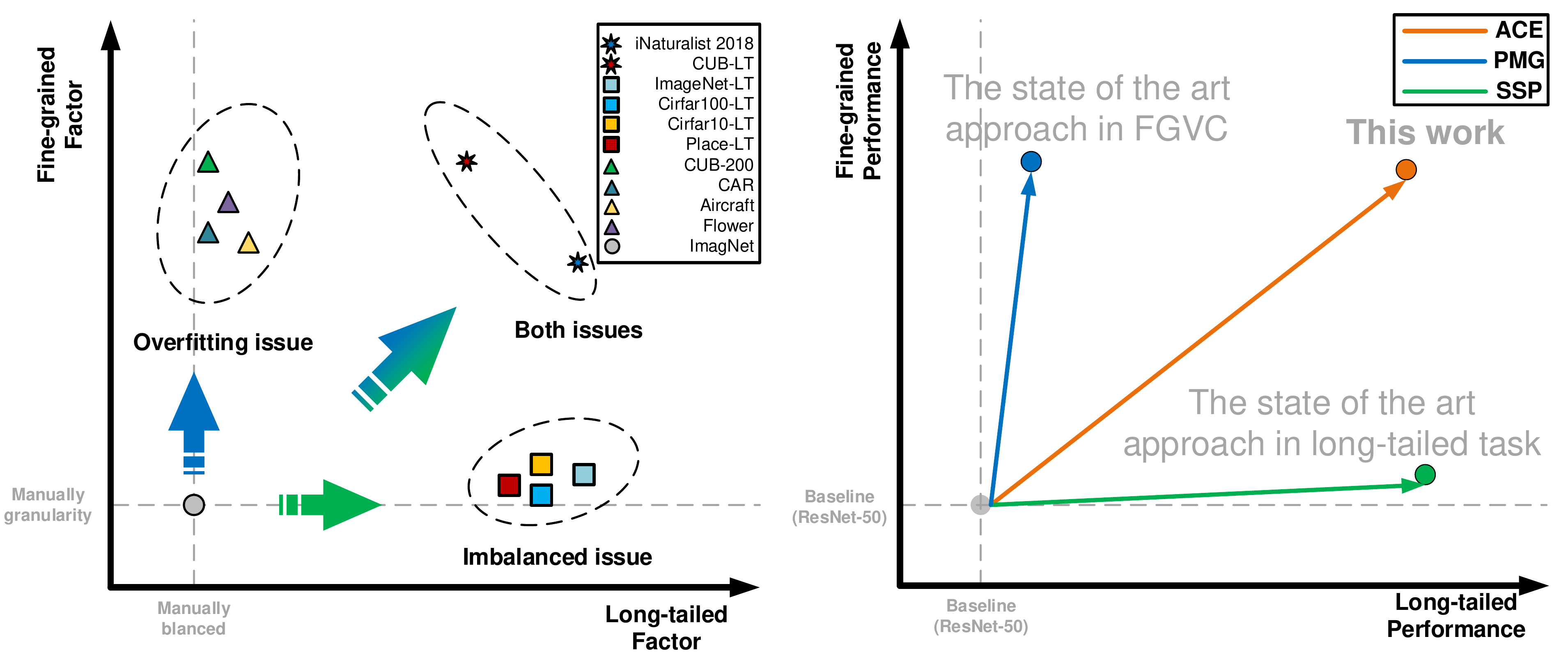}
    \caption{Left: Different datasets have varying degrees of long-tail and fine-grained characteristics. Right: The existing approaches can only solve one aspect of the problem.}
    \label{fig:confusionLoss}
    \vspace{-10pt}
\end{figure*}

Most of the current visual classification tasks usually only have the challenge in one aspect, such as FGVC or long-tailed issues, as mentioned above. However, the data distribution is not always idealistic in the natural world. Comprehensive issues usually accompany it. For instance, Figure~\ref{fig:intro} illustrates two concurrent challenges in the iNaturalist 2018~\citep{van2018inaturalist}. First, the task is a long-tailed distribution with an extremely imbalanced ratio. Since there is a thousand-fold difference in the number of categories, it is hard to learn the tail classes' representation. Meanwhile, it is also an FGVC task where the inter-class similarity and the intra-class variations are subtly intertwined, yielding a daunting classification task, no matter what orders of magnitude of categories (frequent, common, and rare). That makes the model easy to overfit \cite{dubey2018pairwise}.

From the existing literature, there are only a few attempts to solve these two problems simultaneously. Relevant efforts mostly focus on tackling either task. In FGVC, most of the recent research efforts have converged to learn pivotal local/part details relevant to distinguishing fine-grained categories \eg, \cite{fu2017look,yang2018learning,zheng2019looking}, and typically require the fusion of several sophisticated computer vision techniques to accomplish the task such as in \cite{ge2019weakly}. In resolving the long-tailed issue, previous approaches have looked into data balanced sampling \cite{huang2016learning,wang2017learning} and the recent development such as \citet{Kang2020Decoupling} learns the representation at the first stage and refines the classifier by balanced data sampling. Different existing tasks have varying degrees of fine-grained and long-tail factor. As shown in Figure~\ref{fig:confusionLoss} (left), we leverage the maximum imbalanced ratio and the normalized feature cosine similarity between each category as the fine-grained and long-tailed factor to evaluate the characteristic of each task.

Motivated by these developments, we propose a flexible and effective regularization design that aims at guiding the resulting DNN learning to improve model efficiency on tackling the FGVC and long-tailed issues at the same time. Our method is relevant to the {\it pairwise confusion} (PC) \cite{dubey2018pairwise}.  PC only makes an image confuse with another image, but it does not use the rest sufficiently. Furthermore, while PC encounters imbalanced problems, it offers few improvements. The Figure~\ref{fig:confusionLoss} (right) shows that those approaches can alleviate the overfitting issue in FGVC fail in overcoming the imbalanced issue and vice versa. In this paper, the proposed formulation goes beyond the restriction of working on pairs of data and develops a batch norm-based framework with sufficient model capacity to deal with  FGVC and long-tailed issues simultaneously. We first assume all samples/images within a batch are of different classes. A batch-wise matrix norm then models the targeted confusion energy, termed Adaptive Confusion Energy (ACE). The matrix is constructed by including prediction results from all images within a batch and an adaptive matrix to adjust class-specific weights. The former is used to handle the FGVC task or overfitting issue, and the latter is for resolving the long-tailed distribution. To achieve efficient DNN learning, we provide an approximation scheme to ACE so that gradient backpropagation can be readily carried out. The promising experimental results support that ACE has good potential to function as a generic regularizer for solving a wide range of classification tasks, no matter the fine-grained property or imbalanced distribution. 

%
\section{Related Work}
\label{sec:related}
%

Researches in fine-grained and long-tailed visual classification are going on in two different branches. Most articles focus on just one of these issues. We will introduce recent studies on both sides and then briefly explain our approach.
\vspace{-10pt}
\paragraph{FGVC.}
In the early works, the training data are annotated with additional information such as part labels. Along this line, \cite{berg2014birdsnap} explore the labeled part locations to eliminate highly similar object categories for improving the learned classifiers. The approach in \cite{huang2016part} is established based on a two-stream classification network to capture both object-level and part-level information explicitly. However, due to the rapid research advances in visual classification, the most recent FGVC approaches are designed to complete the model learning solely based on category labels' information.
\cite{sun2019fine,dubey2018pairwise,wang2018learning,li2018towards,yang2018learning,zheng2019looking,chen2019destruction, du2020fine}.
\vspace{-10pt}
\paragraph{Long-tailed visual recognition}
To alleviate the impact of the imbalanced data, the two common basic methods are re-sampling and re-weighting. Re-sampling in the early studies includes under-sampling \cite{drummond2003c4} for head categories and over-sampling \cite{ChawlaBHK02,HanWM05,mahajan2018exploring} for tail categories. In recently, the most common strategy is called class-balanced sampling \cite{ShenLH16}. Unlike instance-balanced sampling, every image has the same probability of being selected; class-balanced is to weight the sampling frequency of each image according to the number of samples of different categories. Furthermore, \cite{gupta2019lvis} proposed repeat
factor sampling (RFS), a dynamic-sampling mechanism, to balance the instances. Unlike sampling, because of the flexibility and convenience of loss calculation, many more complex tasks, such as object detection and instance segmentation, are more likely to leverage the re-weighted loss to solve the problem of long-tail distribution. From the reverse weighting based on category distribution to the Hard Example Mining \cite{ShrivastavaGG16} which is carried out directly according to the credibility of classification without knowing the category, such as Focal loss \cite{LinGGHD17} and LDAM \cite{cao2019learning}. Also, due to implementation is easy, some works \cite{CuiJLSB19,JamalB0WG20,TanWLLOYY20} show competitive results in complex tasks.  On the other way, the two-stage training strategies that learn the classifier with re-balancing data and to learn representation with original data is regarded as an effective solution to the constant tail distribution.
\cite{Kang2020Decoupling, zhou2020, LiWKTWLF20, HuJTCMZ20, abs-2007-11978, tang2020long, DBLP:conf/nips/YangX20}.
\vspace{-10pt}
\paragraph{Confusion energy.} 
In FGVC, the confusion-related formulation for dealing with intra-class variations and inter-class similarity has two main implications. First, it can be applied to alleviate the overfitting problem in training an FGVC model. \cite{dubey2018pairwise} construct a Siamese neural network, trained with a loss function including {\em pairwise confusion} (PC). The design reasons that bringing the class probability distributions closer to each other could prevent the learned FGVC model from overfitting sample-specific artifacts. Second, the confusion tactic can be used to boost the FGVC performance by focusing on local evidence. \cite{chen2019destruction} partition each training image into several local regions and then shuffle them by a {\em region confusion mechanism} (RCM). It implicitly excludes the global object structure information and forces the model to predict the category label based on local information. In other words, the ability to identify the object category from local details is expected to be enhanced through shape confusion.

Our approach to FGVC and long-tail is most relevant to the above confusion-based approaches. We retain the advantages of confusion energy and exploit the potential in the long-tailed distribution. And then propose a novel confusion energy term called {\em Adaptive Confusion Energy} (ACE), which can flexibly adjust the confusion strength corresponding to the data distribution.

%
\section{Approach}
\label{sec:approach}
%

We propose the adaptive confusion energy (ACE) to address the image classification on data with natural world distribution. Our ACE module combines two novel components: 1) Batch confusion norm (BCN) and 2) adaptive matrix $A$. We elaborate our method as follows.

\subsection{Overfitting Elimination by Batch Confusion Norm}
\label{sec:BCN}

Given a training set $\cD$ over total $C$ fine-grained categories, an arbitrary sample $\bx$ from $\cD$ is denoted as $(\bI, y)$ where $\bI$ represents an image and $y \in \{1, \dots, C\}$ denotes the corresponding class label. We define a batch $\cB = \{\bx_1, \bx_2, \dots, \bx_M\}$ as a set of sample $\bx_i$ randomly sampled from $\cD$. Note that $M$ is the batch size. For each training sample $\bx_i$ in a batch $\cB$, we forward propagate it through a classification model $\Phi$ and then obtain the predicted probability (\ie, softmax) $\bp_i$. After that we define batch-wise class prediction matrix $P$ by 
\begin{linenomath}
\begin{equation}
    P = \left[\bp_1 \; \bp_2\; \dots\; \bp_M \right] \in \bbR^{C\times M},
\end{equation}
\end{linenomath}
where $\bp_i \in \bbR^{C}$ is the predicted probability over the C fine-grained categories. Notice that, in our BCN module, we assume that $M \le C$ and all images within a batch $\cB$ are randomly sampled from the $\cD$. On the contrary, the confusion regularization of PC \cite{dubey2018pairwise} only affects the paired images with distinct labels. In a nutshell, BCN considers global optimization in substitution a pair.  

The explicit purpose of BCN is to increase the difficulty for a model to learn classification problems by infusing slight classification confusions into the training procedure. To this end, it is reasonable to minimize the rank of the batch-wise class prediction matrix $P$ so that the predictions for all samples in a batch are {\em similar}:
\begin{linenomath}
\begin{equation}
    \argmin_\Phi \mathrm{rank} (P)\,.
    \label{eqn:rank_P}
\end{equation}
\end{linenomath}
However, the rank-related minimization problems are often NP-hard. To address this problem, in this paper, we utilize convex relaxation methods to approximate the solutions. With the help of convex relaxation methods, minimizing the rank of $P$ can be reduced as the minimization of its {\em nuclear norm}. That is the {\em batch confusion norm} of $P$ can be formulated as
\begin{linenomath}
\begin{equation}
    \| P \|_{\mathrm{BCN}} = \| P \|_*
    \label{eqn:norm_BCN}
\end{equation}
\end{linenomath}
where $\|\cdot \|_*$ is the nuclear norm which computes the sum of the singular values of the underlying tensor/matrix.

\paragraph{Stability.} In order to make the matrix decomposition of $P$ stable and prevent the negative singular values from heavily affecting the training loss, we replace the right-hand side of (\ref{eqn:norm_BCN}) with $\| P^\mathsf{T} P \|_*$ since it is known that
\begin{linenomath}
\begin{equation}
    \mathrm{rank}(P) = \mathrm{rank}(P^\mathsf{T} P).
    \label{eqn:rankPP}
\end{equation} 
\end{linenomath}

Finally, by combining all the technique above, our batch confusion norm can be formulated as 
\begin{equation}
    \| P \|_{\mathrm{BCN}} = \| P^\mathsf{T} P \|_*.
    \label{eqn:BCN}
\end{equation} 

\subsection{Data imbalance control via adaptive matrix $A$}
\label{sec:MA}
In this subsection, we introduce the adaptive matrix $A$, which equips the BCN in Sec. \ref{sec:BCN} with the ability to handle imbalanced data and finally evolve into our Adaptive Confusion Energy (ACE). 

Empirically, the classification accuracy of different categories (with various data distribution) depends on different levels of confusion energy. Take the tailed classes with few samples; for example, applying high confusion energy on tailed classes may damage the classification performance. To fix this issue, we adopt an adaptive matrix $A \in \bbR^{C \times C}$ to generalize the BCN. The adaptive matrix $A$ enables the BCN to adjust the strength of confusion energy for each category. Here are a couple of criteria for initializing a proper $A$:
\begin{itemize}
    \item When it comes to a dataset with long-tailed distribution, $A$ should alleviate the confusion energy on the tailed categories to prevent the model from getting excessive confusion over these classes. 
    \item When the data distribution is balanced, $A$ should be approximately the same as an identity matrix.
\end{itemize}

Following these guidelines, we design the adaptive matrix $A$ as
\begin{linenomath}
\begin{align}
    A_{ij} = \left\{ \begin{array}{ll}
    (\frac{\mathcal{N}_i}{\mu})^{\sigma^\tau}, & i = j\\
    0,                          &  i \neq j
    \end{array}
    \right.,
    \label{eqn:A}
\end{align}
\end{linenomath}
where $\mathcal{N}_i, i \in \lbrace 1, 2, ..., C \rbrace$ represents the number of data for each category. Also, $\mu = \frac{1}{C} \sum_{i=1}^C \mathcal{N}_i$ and $\sigma = \sqrt{\frac{1}{C} \sum_{i=1}^C (\mathcal{N}_i - \mu)^2}$ denote the mean and standard deviation of $\mathcal{N}_i$, respectively. Finally, $\tau$ stands for a tunable hyper-parameter. Note that when $\mathcal{N}_i \longrightarrow \mu$, we have $\bA_{ii} \longrightarrow 1$. Also, if $\sigma \longrightarrow 0$ then $\bA_{ii} \longrightarrow 1$. This means that $\bA$ will downgrade to the identity matrix when the data distribution is balanced.

Finally, by incorporating the batch confusion norm with the adaptive matrix $A$, we can now formulate our novel adaptive confusion energy (ACE) as follows. 
\begin{linenomath}
\begin{align}
    \mathcal{L}_\mathrm{ACE} & = \| P^\mathsf{T} A^\mathsf{T} A P \|_*,
    \label{eqn:loss_ACE}
\end{align}
\end{linenomath}
where the adaptive confusion energy loss $\cL_{\mathrm{ACE}}$ is computed based on the eigenvalues of $P^\mathsf{T} A^\mathsf{T} A P$. 
It is worth noting that our ACE has sufficient capability to handle data with natural world distribution by alleviating the overfitting problem in a fine-grained model and considering the imbalance problem in the long-tailed data distribution.

\paragraph{Learnability.} In practice, there is no feasible way to ensure that the parameters of $A$ given in (\ref{eqn:loss_ACE}) is optimal by pre-defined parameters. Therefore, We alternately use it as a {\em good} initialization and set $A$ as a learnable model, denoted as $\hat{A}$. Consequently,  we revise the $\cL_\mathrm{ACE}$ into
\begin{linenomath}
\begin{align}
    \hat{\cL}_\mathrm{ACE} & = \| P^\mathsf{T} \hat{A}^\mathsf{T} \hat{A} P \|_* + \eta \| \hat{A} - A \|_2,
    \label{eqn:learnable}
\end{align}
\end{linenomath}
where $\eta$ is a tunable weight for the regularization term which regulates the learnable adaptive matrix $\hat{A}$ should not be too far away from $A$. In practice, we initialize $\hat{A}$ with $A$ and set $\eta = 1$ to simply improve the original adaptive matrix using the hand-crafted $A$.

\subsection{Loss function}

Combine our ACE in Eq. (\ref{eqn:learnable}) with the original classification loss, the overall objective function can now be easily expressed by
\begin{linenomath}
\begin{equation}
    \cL= \cL_{\mathrm{CE}} + \lambda \, \hat{\cL}_\mathrm{ACE}
    \label{eqn:loss_all}
\end{equation}
\end{linenomath}
\noindent where $\cL_\mathrm{CE}$ is the cross-entropy loss which is usually applied in classification task and $\lambda$ is a hyper-parameter to adjust the influence of the ACE loss to learning the model.

%
\section{Experimental Results}
\label{sec:result}
%

\begin{table*}[ht!]
\caption{Head-to-head comparisons of the confusion energy scenarios on the standard FGVC datasets CUB-200-2011 (CUB), Stanford Cars (Cars), and FGVC-Aircraft (Aircraft).}
\label{tab:fgvc_ab}
\centering
\begin{tabular}{lccccccccccccccc}
\toprule
\multirow{2}{*}{Model} & \multicolumn{3}{c}{ResNet-50} & \multicolumn{3}{c}{ResNeXt-50} & \multicolumn{3}{c}{ResNeXt-101} & \multicolumn{3}{c}{DenseNet-161}\\
        \cmidrule(lr){2-4}
        \cmidrule(lr){5-7}
        \cmidrule(lr){8-10}
        \cmidrule(lr){11-13}
        \cmidrule(lr){14-16}
                    & CUB  & CAR  & AIR  & CUB  & CAR  & AIR  & CUB  & CAR  & AIR & CUB  & CAR  & AIR \\
        \midrule
        \midrule
        Baseline    & 85.5 & 92.7 & 90.3 & 86.3 & 93.1 & 90.9 & 87.3 & 93.5 & 91.6 & 87.5 & 93.4 & 92.7 \\
        PC          & 87.0 & 92.4 & 90.1 & 87.5 & 93.2 & 91.2 & 88.2 & 93.7 & 92.4 & 88.2 & 93.6 & 92.9 \\
        \midrule
        \midrule
        Ours        & \bf 87.8 & \bf 94.3 & \bf 93.2 & \bf 88.1 & \bf 94.4 & \bf 93.3 & \bf 88.6 & \bf 94.5 & \bf 93.5 & \bf 89.2 & \bf 94.8 & \bf 93.5 \\
        \bottomrule
\end{tabular}
\vspace{-5pt}
\end{table*}

\begin{table}[ht!]
\centering
\caption{Compare the results with the typical state-of-the-art. The CNN backbone is ResNet-50.}
\label{tab:fgvc_all}
\resizebox{0.47\textwidth}{!}{
\begin{threeparttable}
\begin{tabular}{lcccc}
    \toprule
    Method & Param. (M) & CUB & CAR & AIR \\
    \midrule
    Baseline & 24 & 85.5 & 92.7 & 90.3 \\
    PC$^\dag$ & 24 & 87.0 & 92.4 & 90.1 \\
    \midrule
    DB      & $\sim$24 & 87.7 & \bf 94.3 & 92.1 \\
    DFL-CNN & $\sim$24 & 87.4 & 93.1 & 91.7 \\
    NTS-Net & $\sim$24 & 87.5 & 93.9 & 91.4 \\
    DCL     & $\sim$24 & 87.8 & 94.5 & 93.0 \\
    iSQRT-COV & $\sim$24 & \bf 88.1 & 92.8 & 90.0 \\
    \midrule
    Ours    & 24 & 87.8 & \bf 94.3 & \bf 93.2 \\
    \midrule
    \midrule
    PC$^\dag$ (DenseNet-161) & 28 & 88.2 & 93.6 & 92.9 \\
    S3N$^\ddag$     & 101 & 88.5 & 94.7 & 92.8 \\
    PMG$^\ddag$     & 45 & 88.9 & 95.0 & 92.8 \\
    \midrule
    Ours (DenseNet-161) & 28 & \bf 89.2 & \bf 94.8 & \bf 93.5 \\
    \bottomrule
\end{tabular}
    \begin{tablenotes}
            \item[$\dag$] Re-implemented by the same training setting as ours.
            \item[$\ddag$] Modified ResNet-50 with additional modules.
    \end{tablenotes}
\end{threeparttable}
}
\vspace{-10pt}
\end{table}

We conduct extensive experiments to evaluate our approach on three balanced benchmark FGVC datasets, imbalanced datasets, and the natural world distribution dataset. We then describe comparisons to prior work as well as the implementation details. We also provide an insightful ablation study for assessing the performance gains of using adaptive confusion energy (ACE). Finally, several visualization examples are demonstrated for further discussions.

\subsection{Datasets}

We first evaluate the effectiveness of the proposed approach on three standard fine-grained visual classification datasets, namely, CUB-200-2011 \cite{WahCUB_200_2011}, Stanford Cars \cite{KrauseStarkDengFei-Fei_3DRR2013}, and FGVC-Aircraft \cite{maji13fine-grained}. 
The data ratio between training and testing sets is about $1:1$ for CUB-200-2011, and Stanford Cars is about $2:1$ in FGVC-Aircraft. The class distribution of the three datasets is nearly balanced, which can be used to measure the proposed method's performance only in the fine-grained scenario with the adaptive matrix $\hat{A}$ approximating identity matrix. Compared with other datasets for the large-scale visual classification task, these three FGVC datasets have fewer training data for each category.

Next, we go through the experiments on the imbalanced datasets, ImageNet-LT \cite{DBLP:journals/corr/abs-1904-05160}. The former is a long-tailed distribution with a low fine-grained factor, confirming whether the proposed approach will adjust on the purely imbalanced dataset. The latter is a fine-grained dataset that also has a long-tailed property. It can more clearly measure the impact of different approaches, \eg, \cite{Kang2020Decoupling,dubey2018pairwise,du2020fine}.

Finally, we then focus on the natural world distribution datasets and CUB-LT \cite{samuel2021generalized} and iNaturalist2018 \cite{van2018inaturalist} 
which has the properties of both fine-grained and long-tailed distribution. Besides, it is also a large-scale dataset. Judging from the recent literature \cite{cao2019learning,Kang2020Decoupling}, this is a reasonably challenging dataset that the performance can serve as an objective measure about our method's usefulness. Finally, we remark that the proposed model does not require any additional annotations in the training process but merely the image-level class annotations.

\subsection{Implementation details}
\label{subsec:details}

We describe the implementation details with FGVC, long-tailed, and the comprehensive task. All our inference results are obtained from end-to-end training except the results on ImageNet-LT. The experimental results are the mean of three run. We implement our method using the Pytorch framework \cite{paszke2017automatic}, and the platform with eight Nvidia V100. The source code will be made available. 
\vspace{-10pt}
\paragraph{FGVC.} Following relevant work \cite{yang2018learning,chen2019destruction,zheng2019looking}, we evaluate our method on the widely-used classification backbone ResNet series \cite{he2016deep} and DenseNet-161 \cite{huang2017densely} which is pre-trained on the ImageNet dataset. For the sake of fair comparison in FGVC training, we use the data augmentation setting as in \citet{chen2019destruction} that the input size is set as  $448 \times 448$, and horizontal flipping is randomly performed. The initial learning rate, the hyper-parameter $\lambda$, and $\tau$ are $0.008$, $10$, and $0$, respectively. The training batch size usually is $16$ if the GPU memory is enough and the training optimizer is Momentum SGD, which accompanies with cosine annealing \cite{loshchilov2016sgdr} as the learning rate decay.
\vspace{-10pt}
\paragraph{Long-tailed visual recognition.} We further evaluate the proposed ACE on the imbalanced datasets, ImageNet-LT. For the sake of fair comparison, we follow the implementation details as in \cite{Kang2020Decoupling} on ImageNet-LT. We present the ResNeXt-50 performance in the following section and the ResNet-10 and ResNeXt-152 at the supplementary. The phenomena between shallow and deep models are almost consistent. Since the ImageNet-LT has a low fine-grained factor but a substantial imbalanced issue, we set the hyper-parameter $\lambda$ and $\tau$ as $0.25$ and $0.1$, respectively.
\vspace{-10pt}
\paragraph{Comprehensive tasks.} Finally, we have experimental results on the CUB-LT and iNaturalist2018. In addition to using similar augmentation schemes, the setting is following \cite{Kang2020Decoupling,cao2019learning}. The backbones are ResNet-50 with 224 $\times$ 224 input size by 90 training epochs in SGD optimization. The batch size is 16, and the initial learning rate is 0.025 with a cosine annealing decreasing schedule. The confusion weight $\lambda$ is $2.0$ and class-wise confusion weight $\tau$ is $0.0$. Moreover, the experiment about CUB-LT is the same as the previous FGVC setting. 
\vspace{-10pt}
\paragraph{Evaluation.} After training on the FGVC, imbalanced, and natural world datasets, we evaluate the models on the corresponding balanced test/validation datasets and report the top-1 accuracy, which is used commonly. The value of accuracy is reported in the format of percentage.

\subsection{Fine-grained}

To investigate the performance of different confusion energies between the different backbones, we conduct an ablation study from shallow to deep on the ResNet-50, ResNeXt-50, ResNeXt101, and DenseNet-161. Table~\ref{tab:fgvc_ab} shows the head-to-head comparison between PC and ACE.  We re-implement the PC at the same training condition, and the experimental results show that ACE has comprehensively improved against PC. Table~\ref{tab:fgvc_all} shows the comparison to the other state-of-the-art approaches with ResNet-50 backbone. Baseline combines with our approach provides a competitive performance. Note that the state-of-the-art PMG \cite{du2020fine} contains four classifiers with ResNet-50 backbone, which leads the size of the parameters becomes 45 million floating points. The size of PMG is larger than the DenseNet-161 backbone, about 29 million. The Table~\ref{tab:fgvc_ab} presents the competitiveness of our approach on the DenseNet-161 against PMG. Moreover, while the recent state-of-the-art PMG meets our ACE, it also improves. However, while the datasets are not large-scale, although ACE gains additional improvement, the confusion energies only provide little help. Hence, look at the FGVC research recently; it seems to have reached the limitation so far. Hence, it is reasonable to go through the more challenging tasks, which are large-scale, fine-grained, and long-tailed.

\subsection{Long-tailed}

\begin{table}[ht!]
    \centering
    \caption{Following the approach \cite{Kang2020Decoupling} on ImageNet-LT, the proposed ACE gains a significant improvement.}
    \label{tab:imagenet_lt}
    \begin{tabular}{lcccc}
        \toprule
        Method   & Many & Median & Few & Total \\
        \midrule
        \midrule
        ResNeXt-50 & 65.9	& 37.5	 & 7.7 & 44.4 \\
        \midrule
        NCM     & 56.6 & 45.3 & 28.1 & 47.3 \\
        cRT     & 61.8 & 46.2 & 27.4 & 49.6 \\
        $\tau$-norm & 59.1 & 46.9 & 30.7 & 49.4 \\
        LWS     & 60.2 & 47.2 & 30.3 & 49.9 \\
        \midrule
        \midrule
        ResNeXt-50 +PC & 63.9 & 35.5 & 8.8 & 42.8 \\
        \midrule
        NCM     & 52.3 & 42.9 & 28.7 & 44.6 \\
        cRT     & 59.3 & 46.1 & 29.5 & 48.9 \\
        LWS     & 57.3 & 46.4 & 29.8 & 48.4 \\
        \midrule
        \midrule
        ResNeXt-50 + ACE & \bf 67.5 & 42.1 & 10.2 & 47.5 \\
        \midrule
        NCM     & 57.9 & 46.7 & 31.0 & 48.9 \\
        cRT     & 63.2 & 48.1 & 29.7 & 51.4 \\
        LWS     & 60.7 & \bf 49.7 & \bf 33.1 & \bf 51.7 \\
        \bottomrule
    \end{tabular}
    \vspace{-10pt}
\end{table}

ImageNet-LT has a low fine-grained factor but contains a strong imbalance issue. It is a reasonable dataset to measure the performance of ACE on the purely long-tailed distribution. Table~\ref{tab:imagenet_lt} shows the experimental results with the strategy same as \cite{Kang2020Decoupling}. At stage 1 with end-to-end training, while the baseline trained with ACE gains a significant improvement, but drop the performance if it is trained with PC. The reason is that PC does not consider the number of each category on the training set, which will destroy the representation learning. ACE has handled the weight of confusion strength to each category, which will carefully alleviate the overfitting issue. Furthermore, through stage 2, no matter cRT or LWS, ACE also gains an additional improvement against baseline or PC. Hence, tackling the imbalanced data distribution with ACE can learn a better representation.

\subsection{Natural World}

\begin{table}[t]
\centering
\caption{The comparison with some recent stat-of-the-art works \cite{Kang2020Decoupling,DBLP:conf/nips/YangX20} on the iNaturalist 2018.}
\label{tab:inat}
\resizebox{0.48\textwidth}{!}{
\begin{threeparttable}
\begin{tabular}{lccccc}
    \toprule
    Method & Backbone & Many & Median & Few & Total \\
    \midrule
    \midrule
    Baseline    & ResNet-50 & \bf 72.2 & 63.0 & 57.2 & 61.7 \\
    PC$^\dag$   & ResNet-50 & 70.9 & 64.6 & 59.6 & 62.1 \\
    \midrule
    LDAM-DRW$^*$        & ResNet-50 & -    & -    & -    & 64.6 \\
    NCM         & ResNet-50 & 55.5 & 57.9 & 59.3 & 58.2 \\
    cRT         & ResNet-50 & 69.0 & 66.0 & 63.2 & 65.2 \\
    LWS         & ResNet-50 & 65.0 & 66.3 & 65.5 & 65.9 \\
    BBN         & ResNet-50 & -    & -    & -    & 66.3 \\
    SSP         & ResNet-50 & -    & -    & -    & 68.1 \\
    \midrule
    Ours        & ResNet-50 & 66.6	& \bf 68.0 & \bf 68.2 & \bf 68.3  \\
    \midrule
    \midrule
    Baseline    & ResNet-152 & \bf 75.2 & 66.3 & 60.7 & 65.0 \\
    PC$^\dag$   & ResNet-152 & 72.1 & 67.2 & 61.3 & 65.9 \\
    \midrule
    NCM         & ResNet-152 & 59.3 & 61.9 & 62.6 & 61.9 \\
    cRT         & ResNet-152 & 73.6 & 69.3 & 66.3 & 68.5  \\
    LWS         & ResNet-152 & 69.4 & 69.5 & 68.6 & 69.1 \\
    \midrule
    Ours        & ResNet-152 & 69.2 & \bf 70.8 & \bf 72.7 & \bf 71.7 \\
    \bottomrule
\end{tabular}
    \begin{tablenotes}
            \item[$\dag$] Re-implemented by the same training setting as ours.
            \item[$*$] The results reproduced with author's code.
    \end{tablenotes}
\end{threeparttable}
}
\vspace{-10pt}
\end{table}

\begin{figure}[ht!]
    \centering
    \includegraphics[width=0.47\textwidth]{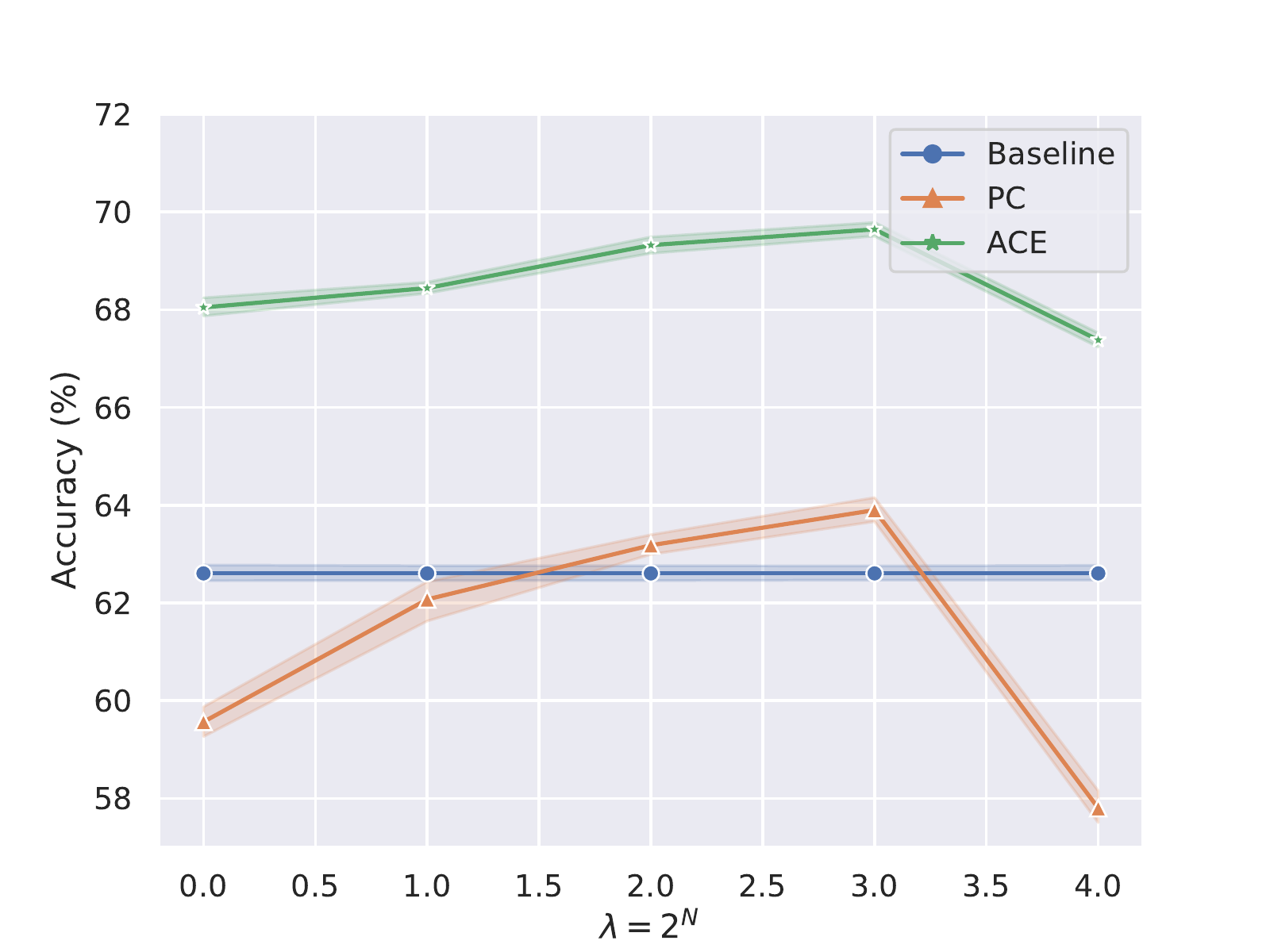}
    \caption{The accuracy with different confusion weight $\lambda$.}
    \label{fig:cub_lt}
    \vspace{-10pt}
\end{figure}

\begin{figure}[ht]
    \centering
    \includegraphics[width=0.47\textwidth]{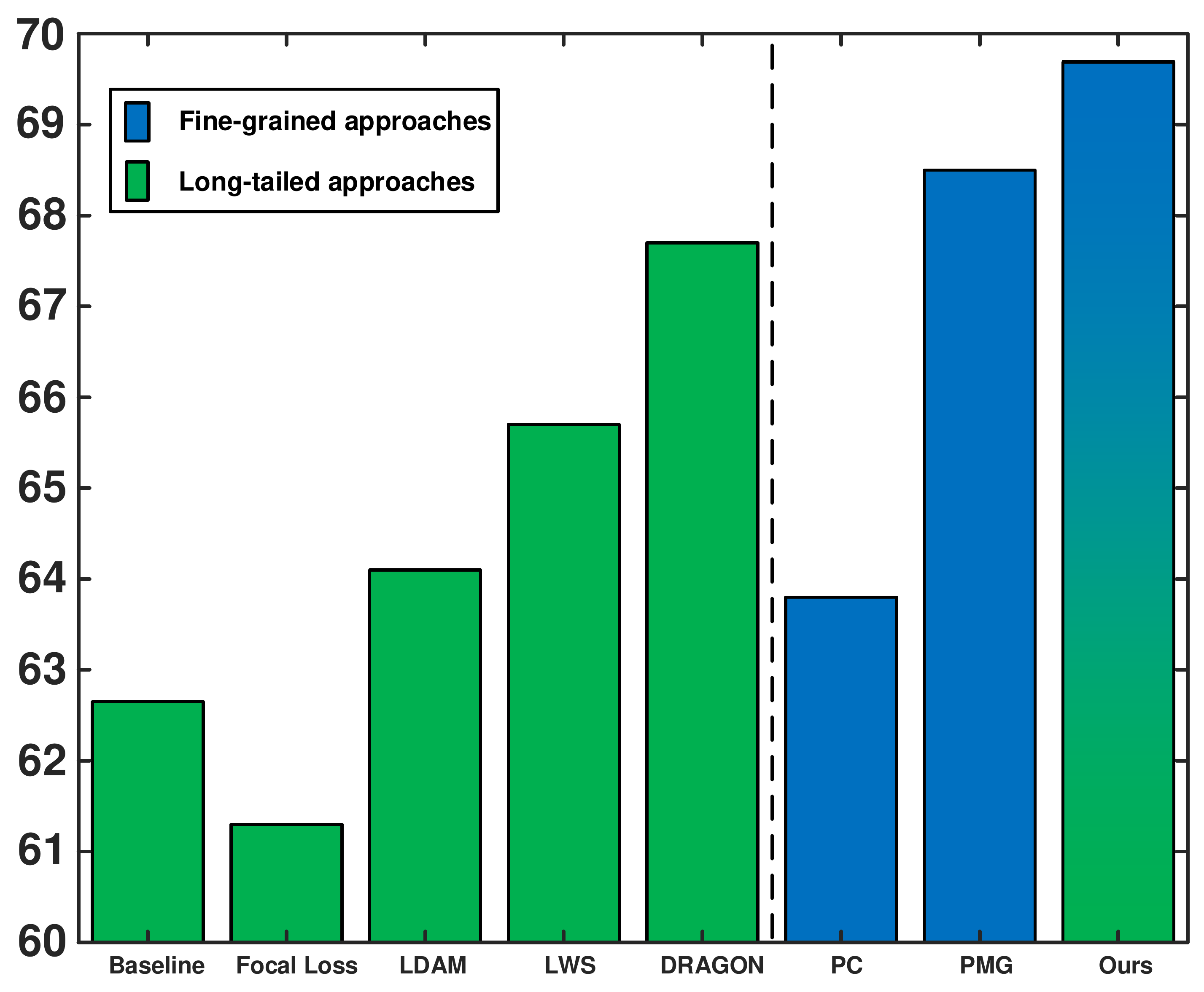}
    \caption{The accuracy of fine-grained and long-tailed approaches in CUB-LT. The proposed work alleviates the concurrent issues.}
    \label{tab:cub_lt}
    \vspace{-10pt}
\end{figure}

Before we look at the nature world dataset, let us quickly look at a small-scale one, CUB-LT, which contains both high fine-grained factors and imbalance issues. It is an ideal dataset for investigating the effect the approaches whether fine-grained \cite{dubey2018pairwise,DBLP:conf/eccv/DuCBXMSG20} or long-tailed \cite{Kang2020Decoupling,samuel2021generalized}. Figure~\ref{fig:cub_lt} shows that PC really can tackle the fine-grained property, but the performance drops when it faces the long-tailed issue. Figure~\ref{tab:cub_lt} presents that the long-tailed approaches only focus on the imbalanced issue but lack the fine-grained property. Similarly, the fine-grained approaches tackle the fine-grained property but not enough to address the imbalanced problem. Hence, the proposed ACE is a comprehensive approach that can easily and efficiently solve fine-grained and imbalanced problems simultaneously.

The proposed ACE preserves the benefits of confusion energy in the FGVC task and addresses the downside of the confusion energy in the long-tailed challenge. Table~\ref{tab:inat} shows the results on the natural world distribution dataset. $\hat{A}$ enables the BCN to focus on the head categories but alleviate the confusion energy effect on the tailed categories. Note that our models are trained not only with the most common way of data sampling {\em instance-balanced sampling} but also end-to-end. In contrast, LWS \cite{Kang2020Decoupling} trains the model in two stages and requires the use of {\em class-balanced sampling}. Besides, SSP \cite{DBLP:conf/nips/YangX20} starts with the self-supervised learning step and then follows the work of \cite{Kang2020Decoupling}, which contains three stages.

\citet{dubey2018pairwise} has shown that confusion energy alleviates the overfitting problem and improves the FGVC performance. However, we observe that if the baseline model is coupled with the confusion energy directly, the overall performance only improves slightly on the natural world dataset. It suggests that the long-tailed issue needs further investigations beyond the model of confusion energy. 

\subsection{Analysis}

\begin{figure*}[t]
    \centering
    \subfloat[Baseline]{\includegraphics[width=0.33\textwidth]{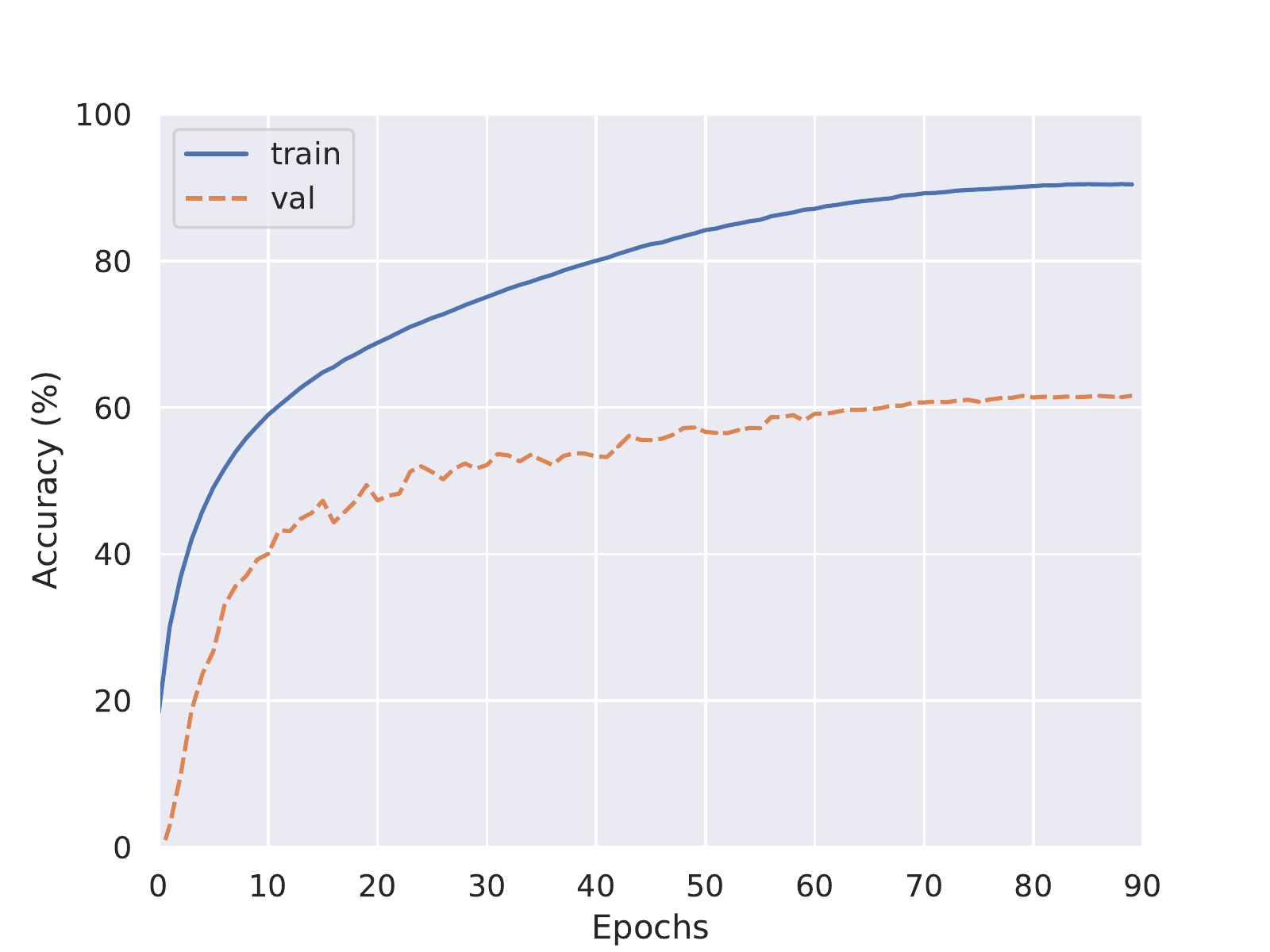}}
    \subfloat[PC]{\includegraphics[width=0.33\textwidth]{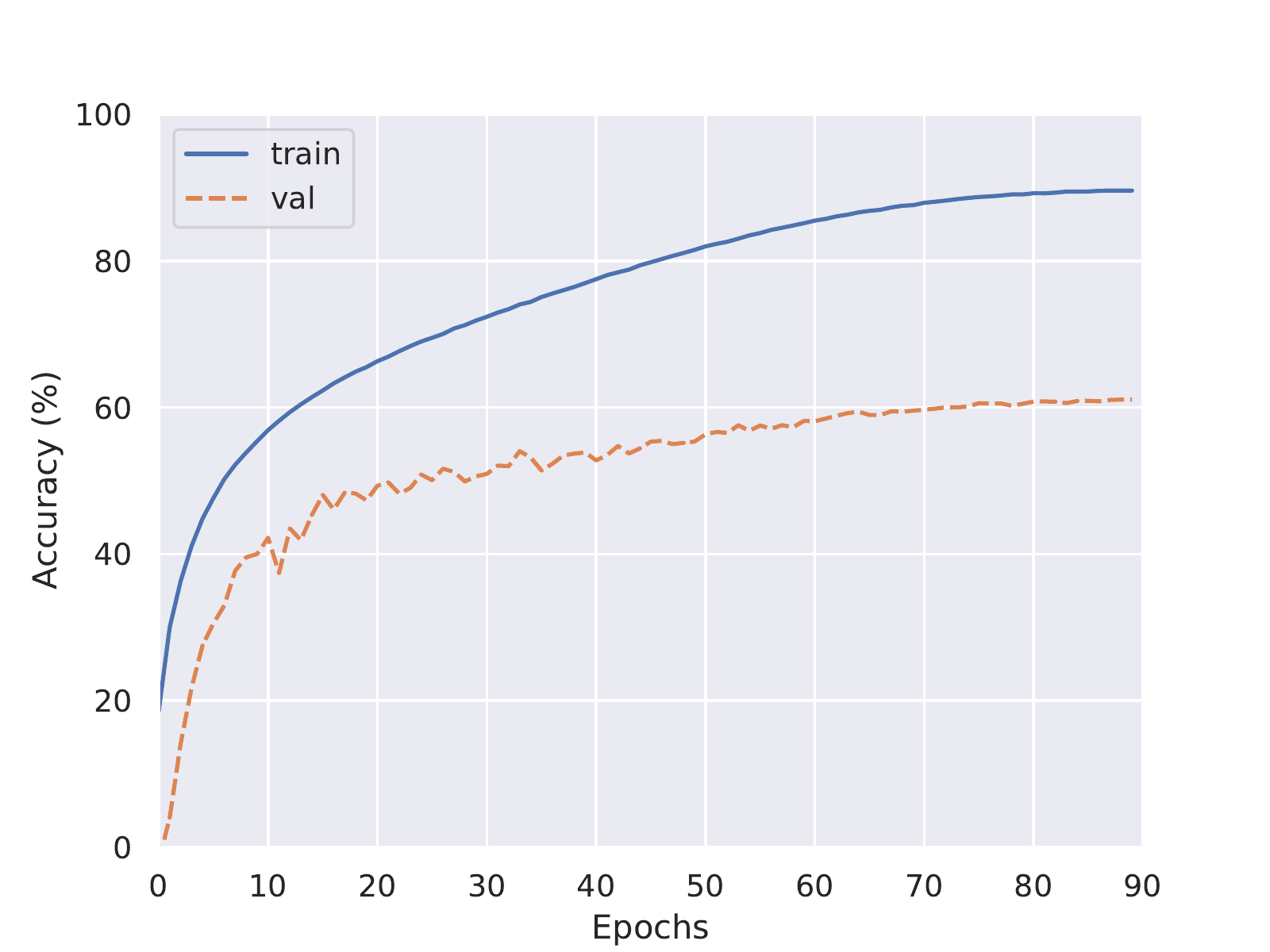}}
    \subfloat[ACE]{\includegraphics[width=0.33\textwidth]{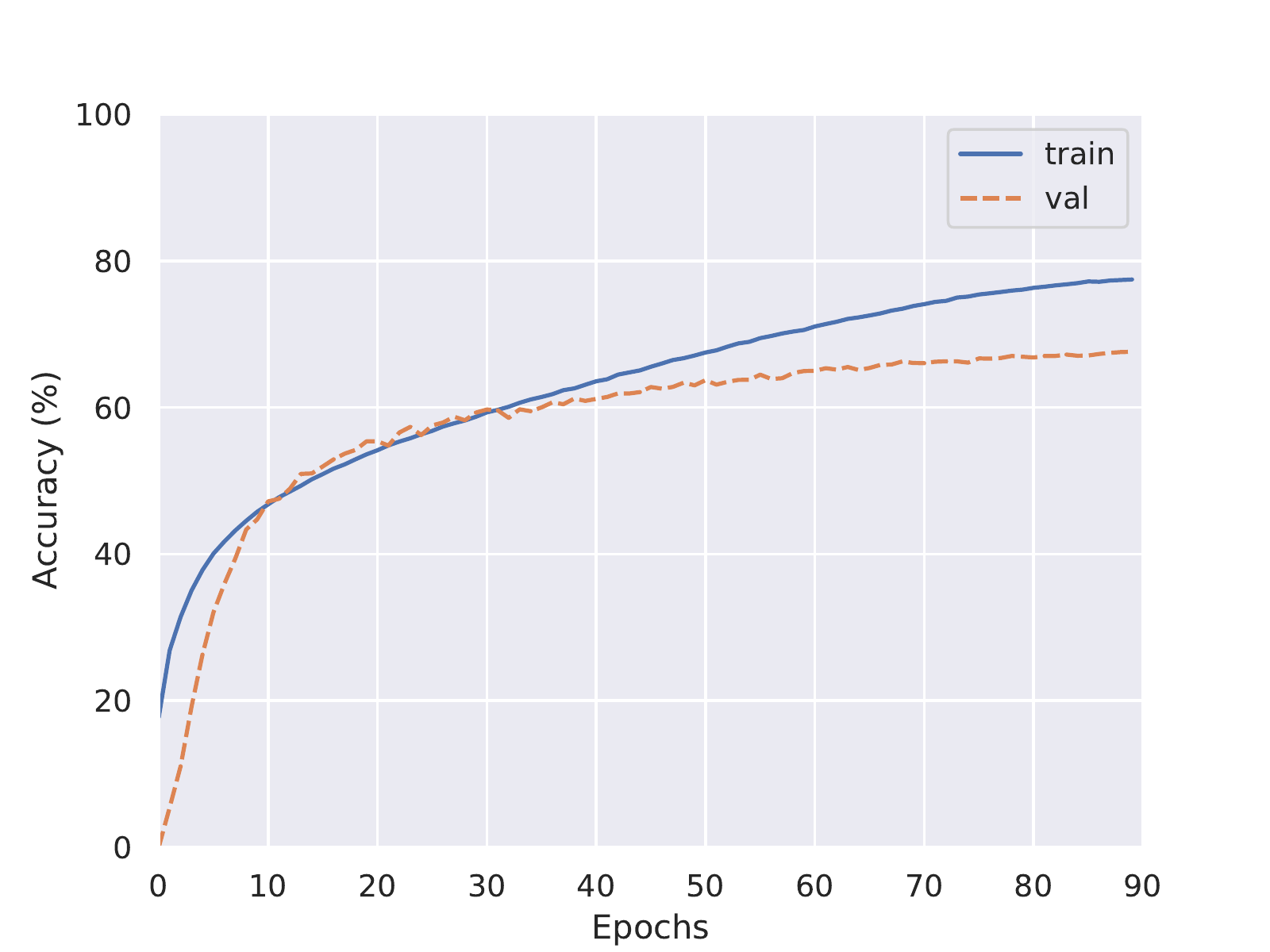}}
    \vspace{-10pt}
    \caption{Observation of the overfitting issue. (a) and (b) shows that there is a large gap between training accuracy and validation performance. (c) presents ACE alleviates the overfitting issue and improves the validation performance.}
    \label{fig:overfit}
    \vspace{-10pt}
\end{figure*}

\begin{figure}[t!]
    \centering
    \includegraphics[width=0.47\textwidth]{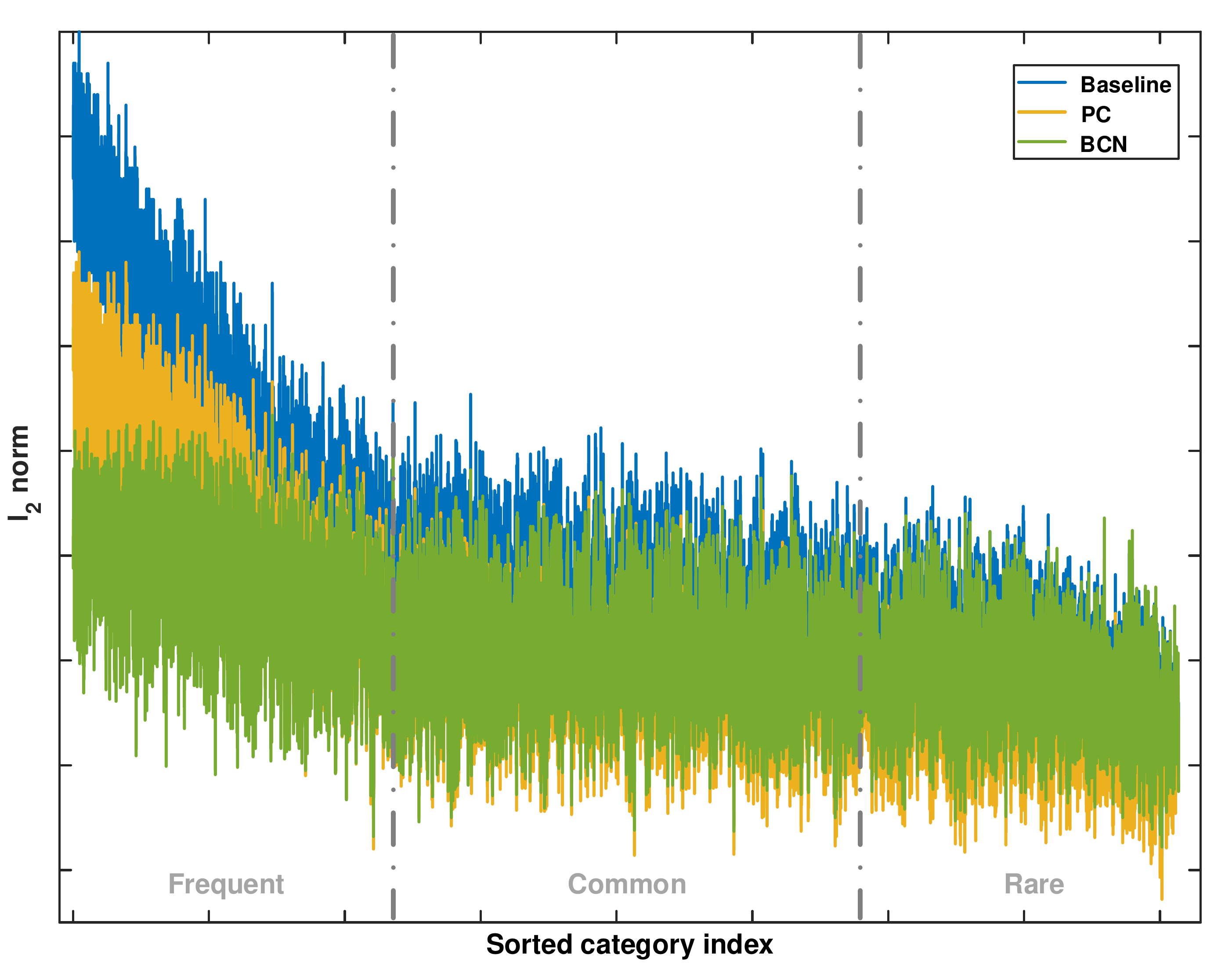}
    \vspace{-15pt}
    \caption{The $l_2$-norm of each category corresponds to the weight $\mathbf{w}_i$ in the classifier.}
    \label{fig:w_norm}
    \vspace{-15pt}
\end{figure}

\begin{figure}[ht!]
    \centering
    \includegraphics[width=0.47\textwidth]{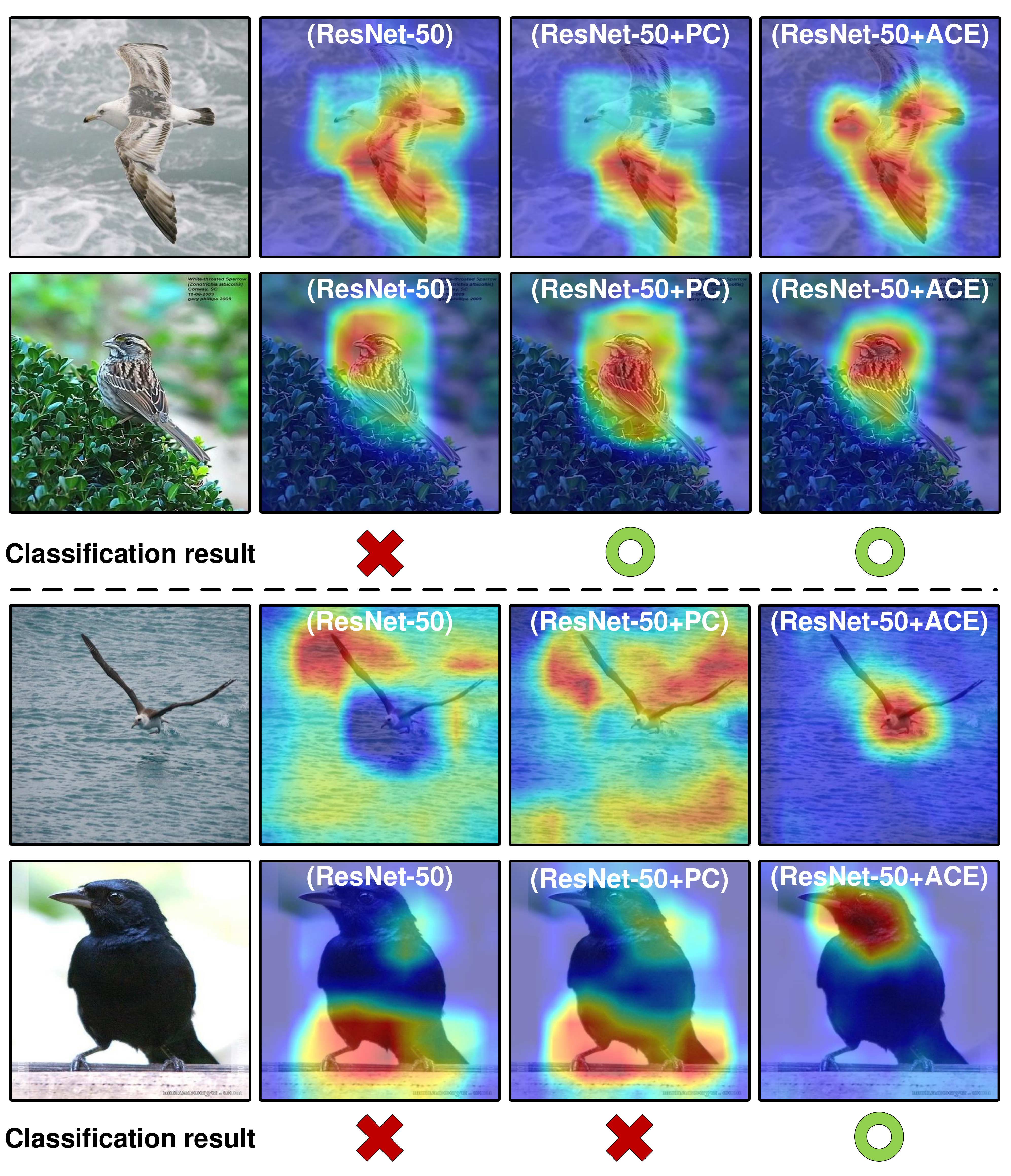}
    \vspace{-15pt}
    \caption{Heatmap-visualization of testing images by Grad-CAM. We show each model's corresponding heatmap. 
    }
    \label{fig:cam}
    \vspace{-15pt}
\end{figure}

In this section, we give some analysis about the influence of ACE. Classification frameworks usually use the cross-entropy loss as the objective function. The training loss always converges to a shallow level, regardless of the characteristics of the dataset. However, this is entirely unreasonable, and to some extent, the model overfits the training data. Hence, we show the performance during training on Figure~\ref{fig:overfit} to present that ACE can prevent the model from overfitting the training data.

Next, consider each category's magnitude corresponding to the classifier weight $\mathbf{w}_i$ in Figure~\ref{fig:w_norm}. The scale of $\| \mathbf{w}_i \|$  distribution on the baseline method is very similar to the data distribution. Although PC has alleviated the scale of the head categories, the distribution does not change significantly.  Nevertheless, the adaptive confusion energy ACE makes the scale of the head to become smoother. This means that the weights of head categories will not dominate the prediction of the classification. 

In summary, ACE provides several benefits. First, it alleviates the overfitting problem of the cross-entropy loss. While training with the cross-entropy loss concerning the ground truth label in the manner of the one-hot vector, the inter-class similarity information is usually significantly suppressed. Consequently, it cannot capture the fine-grained essence by one single cross-entropy loss while handling the overfitting issue. The proposed ACE successfully alleviates this issue. Second, ACE forces the model to learn the inter-class similarity so that the classifier is more focused on the discriminative parts. This phenomenon can be found by using the class activation mapping (Grad-CAM) \cite{selvaraju2017grad} presented in Figure~\ref{fig:cam}. Third, ACE does not require additional processing of inputs and outputs during training. There is no extra cost at inference time, which makes it flexible and applicable to real applications. Finally, ACE solves the confusion energy problem while meets the long-tailed distribution. ACE coupled with the adaptive matrix $\hat{A}$ preserves the benefits of confusion energy in the FGVC task and addresses its downside in the long-tailed scenario.

%
\section{Conclusions}
%

We have developed a general regularization technique specifically designed for addressing the fine-grained visual classification and the long-tailed data distribution problems simultaneously. The proposed adaptive confusion energy (ACE), together with the standard cross-entropy loss, can be used to account for the inherent classification difficulties due to inter-class similarity and intra-class variations. Moreover, it can also solve the long-tailed problem by an adaptive matrix term. The proposed ACE considers the confusion regularization within each training batch and thus is more general than the suitable formulation of pairwise confusion energy. The resulting model can learn discriminative features within regions of interest and alleviate the overfitting problem in training. The provided experimental results nearly achieve state-of-the-art over the three mainstream FGVC datasets and are competitive to leading long-tailed approaches on the imbalanced or natural world distribution datasets. Our future work will focus on generalizing the ACE concept to tensors and extending its applications to other challenging computer vision problems.

\bibliography{reference}
\bibliographystyle{icml2021}

\end{document}